\documentclass[letterpaper,10pt,journal,twoside]{IEEEtran}

\usepackage{amsmath,amsfonts,amssymb}
\usepackage{bm}

\usepackage{algorithm}
\usepackage{algorithmic}

\usepackage{array}
\usepackage{multirow}
\usepackage{booktabs}
\usepackage{tabularx}
\usepackage{float}

\usepackage{graphicx}
\usepackage[justification=centering]{caption}
\usepackage{subcaption}
\usepackage{textcomp}
\usepackage{pifont}
\usepackage{url}
\usepackage{verbatim}
\usepackage{xcolor}
\usepackage{color}
\usepackage{ulem}
\usepackage{soul}
\usepackage{framed}
\usepackage{cite}

\usepackage{hyperref}
\hypersetup{
  colorlinks=true,
  linkcolor=blue,
  citecolor=blue,
  urlcolor=red
}

\graphicspath{{./}{./fig/}{./figures/}{./figure/}{./images/}{./ieeeconf/images/}{./ieeeconf/}{./tex/}}
\makeatletter
\def\input@path{{./}{./tex/}{./table/}{./tables/}{./ieeeconf/table/}{./ieeeconf/}}
\makeatother

\def\BibTeX{{\rm B\kern-.05em{\sc i\kern-.025em b}\kern-.08em
    T\kern-.1667em\lower.7ex\hbox{E}\kern-.125emX}}

\title{Lightweight Learning from Actuation-Space Demonstrations via Flow Matching for Whole-Body Soft Robotic Grasping}

\author{
Liudi~Yang$^{3}$, Yang~Bai$^{4,5}$, Yuhao~Wang$^{1}$,
Ibrahim~Alsarraj$^{1}$,
Gitta~Kutyniok$^{4,5}$,
Zhanchi~Wang$^{2}$,
Ke~Wu$^{1}$%
\thanks{Manuscript received November 12, 2025; Revised February 14, 2026; Accepted March 12, 2026.}%
\thanks{This paper was recommended for publication by Editor Barbara Mazzolai upon evaluation of the Associate Editor and Reviewers comments.}%
\thanks{This work was supported by MBZUAI Projects \#848085.}%
\thanks{$^{1}$Robotics Department, Mohamed bin Zayed University of Artificial Intelligence (MBZUAI), Abu Dhabi, United Arab Emirates.}%
\thanks{$^{2}$Hefei National Research Center for Physical Sciences at the Microscale, University of Science and Technology of China (USTC), Hefei, China.}%
\thanks{$^{3}$University of Freiburg, Freiburg, Germany.}%
\thanks{$^{4}$Ludwig Maximilian University of Munich (LMU Munich), Munich, Germany.}%
\thanks{$^{5}$Munich Center for Machine Learning (MCML), Munich, Germany.}%
\thanks{Corresponding authors: Zhanchi Wang (zhanchi@ustc.edu.cn) and Ke Wu (ke.wu@mbzuai.ac.ae).}%
\thanks{Liudi Yang, Yang Bai, and Yuhao Wang contributed equally to this work.}%
\thanks{Digital Object Identifier (DOI): see top of this page.}%
}
\begin{document}
\maketitle

\begin{abstract}
Robotic grasping under uncertainty remains a fundamental challenge due to its uncertain and contact-rich nature. Traditional rigid robotic hands, with limited degrees of freedom and compliance, rely on complex model-based and heavy feedback controllers to manage such interactions. Soft robots, by contrast, exhibit embodied mechanical intelligence: their underactuated structures and passive flexibility of their whole body naturally accommodate uncertain contacts and enable adaptive behaviors. To harness this capability, we propose a lightweight actuation-space learning framework that infers distributional control representations for whole-body soft robotic grasping directly from deterministic demonstrations using a flow matching model (Rectified Flow), without requiring dense sensing or heavy control loops. Trained with only 30 demonstrations covering less than $8\%$ of the reachable workspace, the learned policy achieved a $97.5\%$ grasp success rate over 1000 trials in simulation. In real-world experiments on 50 uniformly distributed targets, the policy achieved a $100\%$ success rate, generalized to object size variations from $-33\%$ to $+100\%$, and remained stable under execution-time scaling from $20\%$ to $200\%$. These results demonstrate that actuation-space learning effectively embeds mechanical intelligence into control, significantly reducing reliance on centralized computation for grasping under uncertainty.
\end{abstract}


\section{Introduction}

Robotic grasping in unstructured environments is inherently contact-rich and uncertain, making it difficult to model and control reliably~\cite{mason2018toward}. 
Rigid hands, with low compliance, often exhibit limited adaptability and safety under such uncertainty. 
Soft robots instead exploit whole-body compliance and redundancy to passively absorb disturbances and tolerate contact variations~\cite{chu2023full}, reducing reliance on precise centralized control, often described as embodied/mechanical intelligence~\cite{mengaldo2022concise}. A large body of soft grasping and manipulation research has focused on tendon-driven soft grippers/hands and cable-driven continuum manipulators, which enable adaptive interaction through compliance and underactuated actuation~\cite{rao2021model}.
Yet planning and control remain challenging because soft-body dynamics are high-dimensional and history-dependent, and friction, material fatigue, and geometric constraints can dominate behavior during interaction~\cite{wang2025spirobs}.

Existing approaches on the control
of soft robots can be broadly categorized into model-based~\cite{della2023model} and learning-based methods~\cite{chen2024data}.
Model-based control relies on explicit physical formulations, ranging from piecewise constant curvature (PCC) models~\cite{chirikjian1992theory} and pseudo–rigid-body approximations~\cite{wang2025spirobs}, to geometrically nonlinear beam and rod theories~\cite{tummers2023cosserat}, and finite-element or solid-mechanics-based approaches~\cite{duriez2013control}.
These controllers offer interpretable descriptions of soft-body mechanics and can often guarantee stability or convergence.
However, their practical deployment is hindered by high modeling complexity~\cite{tummers2023cosserat}, computationally expensive inverse solutions~\cite{duriez2013control}, and large sim-to-real discrepancies arising from material and contact uncertainties~\cite{chirikjian2015conformational}.
Consequently, purely model-based control remains difficult to scale and lacks robustness when dealing with unmodeled deformations or uncertain interactions.

\begin{figure*}[t]
	\vspace*{0.3cm} 
    \centering
    \includegraphics[trim=0 0 0 0, clip, scale=0.9]{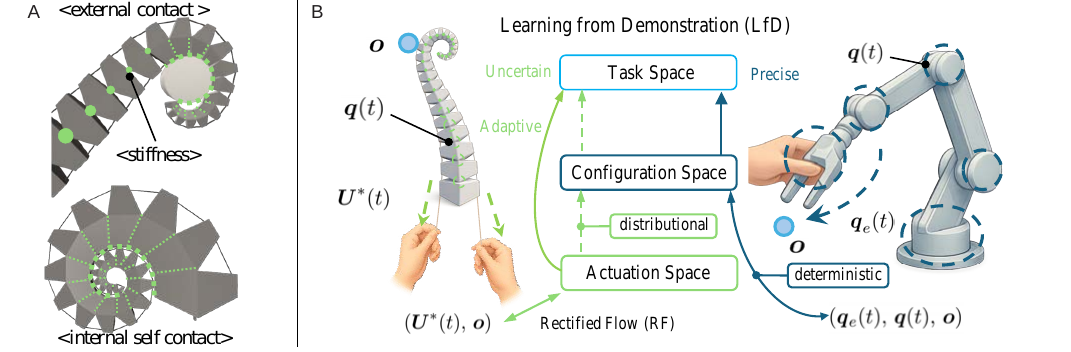}
    \caption{Learning from Actuation-Space Demonstration for Grasping.  
    A. SpiRob. B. Distinction in LfD schemes between rigid and soft robots.}
    \label{comparison}
\end{figure*}

In contrast, learning-based control circumvents explicit modeling by leveraging data~\cite{chen2024data}.
Artificial neural networks (ANNs) have been widely adopted in soft robotics for tasks such as kinematics, dynamics, and state estimation~\cite{george2018control}.
For instance, ANNs have been used to approximate the inverse statics of cable-driven soft arms~\cite{giorelli2015neural}, to model forward dynamics for predictive control~\cite{thuruthel2018model}, and to implement imitation learning policies for soft robotic hands trained from human demonstrations~\cite{gupta2016learning}.
With their universal function approximation capability~\cite{lecun2015deep}, ANNs can capture nonlinear dynamics directly from data without explicit physical models. However, their accuracy and generalization strongly depend on the amount and quality of training data, and limited demonstrations often constrain their robustness and adaptability \cite{ebert2021bridge}.
While this suffices for low-level tasks~\cite{chen2024data}, these methods generally fail to cope with {robotic manipulation scenarios} involving strong dynamics and significant uncertainty~\cite{osa2018algorithmic}. To move beyond supervised imitation, reinforcement learning (RL) has been explored as an alternative that enables autonomous exploration and policy optimization~\cite{kober2013reinforcement}.
In soft robotics, reinforcement learning (RL) has been mainly demonstrated in locomotion and basic trajectory tracking, such as soft snake robots~\cite{liu2023reinforcement} and continuum arms~\cite{thuruthel2018model,satheeshbabu2019open}.
Recent works have begun to explore interaction-rich manipulation with soft robotic arms, including pushing behaviors~\cite{alessi2024pushing} and RL-based characterization of continuum arms~\cite{morimoto2022characterization}.
Nevertheless, applications of RL to whole-body grasping with strong generalization under uncertainty remain scarce~\cite{george2018control}.

Although RL can theoretically surpass expert demonstrations, it is notoriously sample inefficient, requires carefully engineered rewards, and poses safety risks on hardware~\cite{kober2013reinforcement}.
Learning from demonstrations (LfD) reduces the reliance on costly exploration~\cite{nair2018overcoming}.
However, for highly redundant soft robots, the key challenge shifts to how generalization can be achieved from a limited set of demonstrations~\cite{chen2024data}.
More recently, Diffusion Policy (DP) has achieved impressive performance in rigid robot visuomotor control by modeling multimodal action distributions and handling high-dimensional spaces~\cite{chi2025diffusion}.
Extensions include on-device diffusion transformers~\cite{wu2025device}, hierarchical policies for multi-task manipulation~\cite{ma2024hierarchical}, and 3D visual representations~\cite{ze20243d}, further enhancing their generalization across tasks and environments. Despite their success, diffusion-based methods require multi-step denoising at inference, introducing latency and significant computational cost~\cite{chi2025diffusion}.
Such precise, high-frequency closed-loop corrections may not align with the embodied nature of soft robots, where redundancy and compliance inherently mitigate uncertainty~\cite{mengaldo2022concise}, making heavy per-timestep state observation and diffusion-based inference computationally inefficient.

These observations motivate a lightweight, diffusion-inspired flow-matching approach that generalizes from limited demonstrations while avoiding heavy per-timestep inference.
Specifically, we employ the Rectified Flow (RF) model~\cite{rflow}, a flow-matching generative framework that efficiently learns distributional control policies from minimal demonstrations under task conditions~\cite{flowmatching}.
We validate the proposed approach on a spiral-shaped soft robot (SpiRob)~\cite{wang2025spirobs} for whole-body grasping under uncertainty.
Trained with only 30 demonstrations covering less than $8\%$ of the reachable workspace, the learned policy achieves a $97.5\%$ grasp success rate over 1000 simulation trials. In real-world experiments, 50 uniformly distributed targets demonstrate a $100\%$ grasp success rate, confirming robust adaptability under uncertainty. The approach further generalizes to object size variations ranging from $-33\%$ to $+100\%$, extends to diverse object shapes, and remains stable when the execution duration is scaled between $20\%$ and $200\%$. The main contributions of this work are summarized as follows:

\begin{itemize}
\item \textbf{Framework.} A lightweight framework for \textbf{Learning from actuation-space demonstrations} that couples the \textbf{embodied mechanical intelligence} of soft robots with distributional learning to handle uncertainty.
\item \textbf{Method.} A concrete implementation using the \textbf{Rectified Flow} model, enabling efficient distributional control through flow matching with minimal sampling.
  \item \textbf{Evidence.} Comprehensive validation on a SpiRob showing \textbf{strong generalization and robustness} under uncertainty.
\item \textbf{Insight.} A broader perspective that couples \textbf{embodied mechanical intelligence} with \textbf{distributional control policies}, \textbf{alleviating centralized computation} and enabling reliable whole-body grasping under uncertainty.
\end{itemize}

\section{Problem Statement}

\subsection{Task Objectives}

We study whole-body grasping in a horizontal plane using a soft continuum arm. 
Given a cylindrical object at a specified planar position, the robot must approach the target and achieve enclosure by uncurling and then curling its body. 
For repeatable evaluation and to isolate grasp synthesis from uncontrolled object sliding/rolling, the object is constrained (fixed) during the wrapping phase; this also reflects supported or handover-like settings where the object is externally stabilized.

\subsection{Investigated Soft Robot – SpiRob}
Spiral robots (SpiRobs) are a class of bio-inspired continuum manipulators that exploit the geometric regularity of logarithmic spirals~\cite{wang2025spirobs}.
Unlike many tendon-/cable-driven continuum robots that obtain dexterity through multi-tendon and/or multi-segment actuation~\cite{rao2021model}, SpiRobs achieve a wide workspace and smooth transitions among motion primitives such as reaching, wrapping, and grasping with only two tendons on a logarithmic-spiral body (Fig.~\ref{comparison}A). This minimalist design reduces structural complexity and facilitates rapid fabrication (e.g., 3D-printed TPU), making SpiRobs an attractive platform for whole-body grasping. However, the same minimal actuation also introduces substantial control challenges. 
Due to strong underactuation and distributed contact, friction and hysteresis can significantly alter global deformation and interaction forces, making the actuation-to-interaction mapping highly nonlinear and history-dependent~\cite{wang2024exploiting}. This underactuated, contact-rich complexity motivates learning coordinated actuation patterns directly from demonstrations.

\section{Methodology}
\subsection{Preliminary}

\begin{figure}[t]
	\vspace*{0.3cm} 
    \centering  \includegraphics[width=0.85\linewidth]{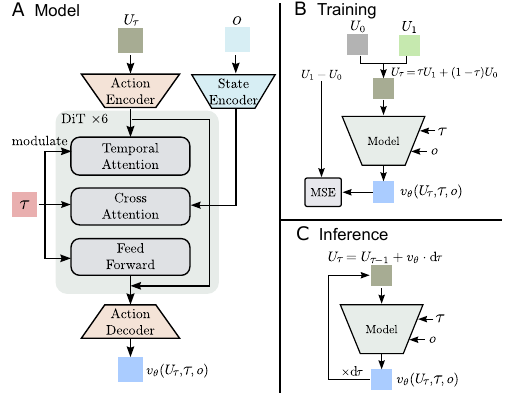}
    \caption{Illustration of the proposed framework. A. Overview of the model architecture designed to learn the flow. B. Training scheme used for optimization. C. Inference scheme applied at test time.}
    \label{fig:model_architecture}
\end{figure}

\subsubsection{Rectified Flow}
Rectified Flow (RF)~\cite{rflow,flowmatching} is a generative flow-matching framework that learns a continuous flow field to transform noise into data, offering a fast and stable alternative to diffusion-based training~\cite{ddpm}.
The overall training and inference pipeline of the RF model is illustrated in Fig.~\ref{fig:model_architecture}. Let $U_1 \sim \Pi_{\text{data}}$ denote an expert training sample, and $U_0 \sim \mathcal{N}(0,\mathbb{I})$ a Gaussian noise sample. 
RF learns a vector field $v_\tau(U_\tau)$ that continuously transports $U_0$ to $U_1$ over the normalized flow time $\tau \in [0,1]$, governed by the ordinary differential equation:
\begin{equation}
\small
    \frac{\mathrm{d}U_\tau}{\mathrm{d}\tau} = v_\tau(U_\tau), \ \tau \in [0,1]
\end{equation}
A simple and natural target for this flow is the straight-line interpolation between $U_0$ and $U_1$:
\begin{equation}
\small
\label{eqn:flow}
     \frac{\mathrm{d}U_\tau}{\mathrm{d}\tau} = U_1 - U_0
\end{equation}
Here, $(U_1 - U_0)$ represents the target action flow, which means the ideal transport direction from noise to data in the action space. RF parameterizes the flow field using a neural network $v_\theta(U_\tau, \tau, o)$, 
where $\theta$ are learnable parameters and $o$ denotes task-specific conditions. 
The model is optimized using the mean squared error (MSE) between the predicted and target action flows:
\begin{equation}
\small
    L = \int_0^1 \mathbb{E}_{U_0, U_1} 
    \left[ \| (U_1-U_0) - v_{\theta}(U_\tau, \tau, o) \|_2^2 \right]
    \mathrm{d}\tau
\end{equation}
where $\mathbb{E}$ denotes the expectation over sampled pairs $(U_0,U_1)$, 
and $\|\cdot\|_2^2$ denotes the squared Euclidean distance, 
computed as the sum of squared element-wise differences.
After training, the learned flow field $v_\theta$ is integrated from $\tau=0$ to $\tau=1$, 
starting from a noise sample $U_0$, to generate a new expert-like sequence $U_1$.
Note that the flow time $\tau$ is a normalized variable for learning the transport process 
and is independent of the robot’s physical execution time.
\subsubsection{Function Representation and Discrete Sampling}\label{representation}

Let $X \in \mathbb{R}^{N \times d}$ denote a discrete trajectory consisting of $N$ sampled timesteps within the time interval $[0, T]$.  
The interpolation operator $\mathcal{I}[\cdot]$, defined through least-squares curve fitting, maps the discrete samples to a continuous vector function $X^*(t) \in \mathbb{R}^{1 \times d}$, which is expressed as
\begin{equation}
\small
    X^*(t) = \mathcal{I}[X(t)], \  t \in [0, T]
    \label{eq:interpolation}
\end{equation}
Conversely, the inverse operator $\mathcal{I}^{-1}[\cdot]$ uniformly samples $N$ points from the continuous vector function $X^*(t)$, giving the discrete representation:
\begin{equation}
\small
    X(t) = \mathcal{I}^{-1}[X^{*}] = [X^{*}(t_1),X^{*}(t_2),X^{*}(t_3)...X^{*}(t_N)]^{\mathrm{T}}
    \label{eq:inverse_interpolation}
\end{equation}

\subsection{Training–Inference Framework}

\subsubsection{Learning from Actuation-Space Demonstration}
\label{actuation_space_learning}
\textbf{Soft Robot Dynamics.}  
Soft robots are inherently underactuated systems, characterized by a continuous backbone with a high-dimensional configuration space but only a limited number of actuators~\cite{chirikjian1992theory}.
The studied SpiRob (Fig.~\ref{comparison}A) is modeled in MuJoCo using a pseudo-rigid-body formulation~\cite{howell1996loop}, where the continuum structure is discretized into a chain of rigid segments connected by revolute joints.
Its dynamics are expressed as
\begin{equation}
\small
M(q)\ddot{q} + C(q,\dot{q}) + G(q) + K(q) + D(q,\dot{q})
= H_c^{\mathrm{T}}(q)U^* + H_e^{\mathrm{T}}(q)u_e,
\label{soft_dynamics}
\end{equation}
where $q$, $\dot{q}$, and $\ddot{q}\in\mathbb{R}^{n}$ denote the generalized displacement, velocity, and acceleration, respectively.
Here, $M(q)$ denotes the inertia matrix, $C(q,\dot{q})$ captures Coriolis and centrifugal effects, $G(q)$ represents gravity, $K(q)$ models the elastic stiffness of the backbone, and $D(q,\dot{q})$ aggregates internal damping and frictional effects.
External and self-contact forces are represented by $H_e^{\mathrm{T}}(q)u_e$. Crucially, the actuation mapping matrix $H_c(q)\in\mathbb{R}^{m\times n}$ is rectangular with $m<n$, reflecting strong underactuation.
For the investigated SpiRob, only $m=2$ tendons are available to actuate a system with $n=24$ degrees of freedom.
Consequently, the set of configurations reachable through actuation constitutes only a subset of the configuration space \cite{brockett1983asymptotic}.
As a result, a directly specified or demonstrated configuration trajectory $q(t)$ may admit no corresponding actuation input $U^*(t)$ capable of realizing it.
Therefore, for underactuated soft robots, configuration trajectories $q(t)$ are not valid control variables for demonstration or inversion, but instead emerge as outcomes of actuation, system dynamics, and interactions with the environment.

\noindent\textbf{Rigid Robot Dynamics.}  
In contrast, rigid robots are typically fully actuated systems, where the generalized coordinates $q(t)$ and the actuation input $U^*(t)$ are well defined through joint-space feedback on $q$:
\begin{equation}
\small
M(q)\ddot{q} + C(q, \dot{q}) + G(q) + D(q, \dot{q})
= H_c^{\mathrm{T}}(q) U^* + H_e^{\mathrm{T}}(q)u_e,
\label{rigid_dynamics}
\end{equation}
with $H_c(q)\in\mathbb{R}^{n\times n}$ being a square, full-rank matrix.
This full-rank actuation mapping allows demonstrations to be naturally specified in the configuration space.
During demonstration, dragging the end-effector along a desired trajectory $q_e(t)$ uniquely determines the joint configuration $q(t)$, which is tracked by a joint-space displacement feedback controller to generate the corresponding actuation input $U^*(t)$, resulting in deterministic and repeatable control (Fig.~\ref{comparison}B).

\noindent\textbf{Comparison and Motivation.}  
As discussed above, configuration-space demonstrations, which are natural and effective for fully actuated rigid robots, do not directly transfer to soft robots.
Due to inherent underactuation and redundancy, soft robots cannot be reliably controlled or demonstrated through configuration trajectories. Instead, the actuation space provides a more appropriate interface for demonstration, as it directly captures the robot’s embodied mechanical intelligence arising from compliance and redundancy.
Learning from demonstrations in the actuation space therefore leverages these intrinsic properties, enabling adaptive behaviors without requiring explicit inverse mappings from configuration to actuation. Accordingly, demonstrations for soft robots are represented as actuation--task pairs
\begin{equation}
\small
(U^*(t),\, o),
\end{equation}
where $U^*(t)$ denotes the demonstrated control sequence and $o$ represents the task condition (Cartesian coordinates of the object to be grasped).
Training in the actuation space allows the robot to learn the correspondence between actuation inputs and task outcomes, supporting robust adaptation in contact-rich and uncertain environments (Fig.~\ref{comparison}B).

\subsubsection{Model Architecture}

To learn the control strategy of SpiRobs in whole-body grasping under uncertainty via RF,  
we develop a MuJoCo-based training framework forming the foundation of our model architecture.  
As illustrated in Fig.~\ref{fig:model_architecture}A, the proposed rectified-flow model, conditioned on the task condition (the object state $o \in \mathbb{R}^2$ in this studied case), learns an approximated flow field $v_{\theta}(U_\tau, \tau, o)$ that transforms Gaussian noise $U_0 \in \mathbb{R}^{N \times 2}$ into the expert control sequence $U_1 \in \mathbb{R}^{N \times 2}$ for the studied SpiRob, where $N$ is the number of timesteps. During execution, the trained policy performs a single open-loop inference at the beginning of the grasp, taking $o$ as input and generating the the expert control sequence $U_1$ without online replanning.


The input action interpolation $U_\tau = (1-\tau)U_0 + \tau U_1$  
is processed by the Action Encoder, a stack of linear layers mapping each interpolated action to a latent feature representation.  
A Diffusion Transformer (DiT)~\cite{dit} backbone learns the rectified flow vector field,  
with Temporal Attention capturing temporal dependencies  
and Cross Attention integrating encoded object-state features for precise spatial conditioning ~\cite{attention}.  
The State Encoder employs linear projections to transform raw object states into compact latent representations suitable for attention-based fusion.  

Each expert demonstration, represented as a continuous control trajectory $U_1^{*}(t)$,  
is discretized using the sampling operator $\mathcal{I}^{-1}[\cdot]$  
(defined in Sec.~\ref{representation}, Eq.~\ref{eq:inverse_interpolation}),  
yielding $U_1 = \mathcal{I}^{-1}[U_1^{*}(t)] \in \mathbb{R}^{N\times 2}$,  
which serves as the target reference during training.  

For each normalized flow time $\tau \in [0,1]$,  
the model generates scale, shift, and gate factors 
to modulate intermediate feature representations within transformer blocks:
$
\small
    h' = \gamma \odot (\alpha \odot h + \beta)
$
where $\alpha$, $\beta$, and $\gamma$ denote the scale, shift, and gate factors, respectively.  
This modulation mechanism adaptively regulates feature magnitudes  
and ensures consistent evolution of the internal flow representation along $\tau$.  
An Action Decoder, implemented as a convolutional neural network (CNN),  
maps the latent features back to the target control sequence space.  

During training (Fig.~\ref{fig:model_architecture}B),  
the interpolated action $U_\tau = (1-\tau)U_0 + \tau U_1$ is fed into the model,  
which is optimized by minimizing the mean squared error (MSE)  
between the predicted and target action flows $(U_1 - U_0)$.  
During inference (Fig.~\ref{fig:model_architecture}C),   
the model starts from $U_0$ and iteratively integrates  
the learned flow field $v_{\theta}(U_\tau, \tau, o)$ using the Euler method  
to reconstruct the final discrete sequence $\hat{U}_1$(where the hat symbol $(\hat{\cdot})$ indicates model-inferred quantities).  
Finally, the continuous control function for execution is obtained as  
$\hat{U}_1^{*}(t) = \mathcal{I}[\hat{U}_1]$  
(using the interpolation operator $\mathcal{I}[\cdot]$ defined in Sec.~\ref{representation}, Eq.~\ref{eq:interpolation})  
to recover temporal continuity for real-robot deployment.

\section{Learning in Simulation}

\begin{figure}[t]
	\vspace*{0.3cm} 
    \centering
    \includegraphics[trim=0 0 0 0, clip, scale=1.1]{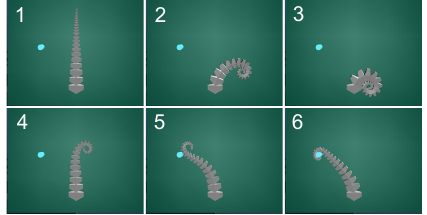}
    \caption{Example of an expert grasping demo in the simulation.}
    \label{fig:simulation_data}
\end{figure}

\subsection{Data Collection in Simulation}

In simulation, the manipulated object is fixed on a planar surface with its state $o \in \mathbb{R}^2$ representing the object's 2D Cartesian coordinates. The SpiRob is aligned parallel to the tabletop. 
The object is modeled as a cylinder of 30\,mm diameter,  and placed on the table, as shown in Fig.~\ref{fig:simulation_data}.
Physical parameters such as mass, stiffness, and damping follow the settings in~\cite{wang2025spirobs}.

\subsubsection{Training Region within the Workspace}

\begin{figure}[t]
    \centering
    \includegraphics[trim=0 0 0 0, clip, scale=0.50]{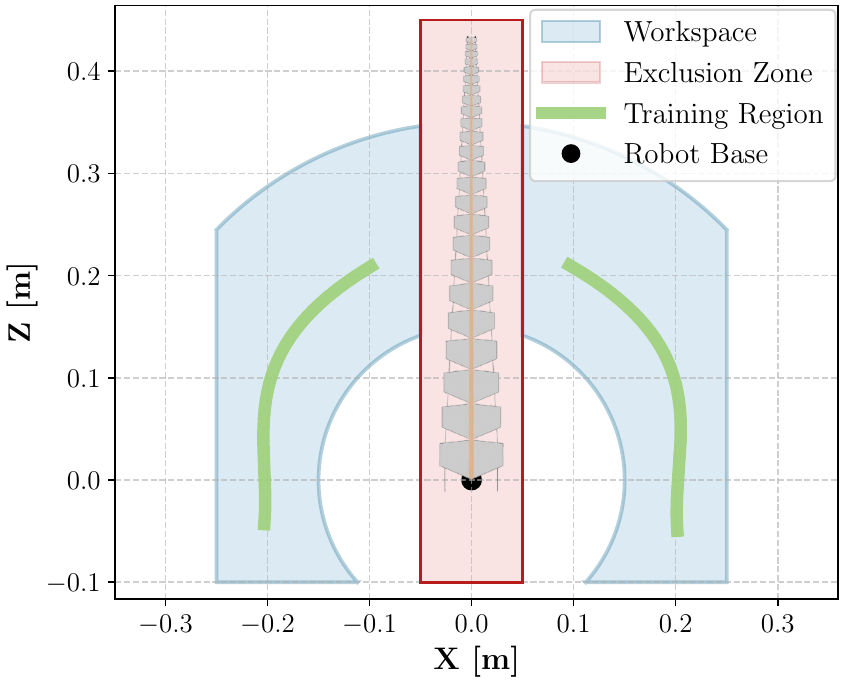}
    \caption{Workspace and training region configuration for data generation.}
    \label{fig:workspace}
\end{figure}

The overall workspace $\mathcal{R}_{\mathrm{val}}$ is defined as in Fig.~\ref{fig:workspace}:  
\begin{equation}
\small
\begin{aligned}
&\mathcal{R}_{\mathrm{val}} = 
\Bigl\{(x,z) \;\big|\;
r_{\mathrm{in}} \leq \sqrt{(x-x_0)^2 + (z-z_0)^2} \leq r_{\mathrm{out}} \Bigr\} \\
&\;\cap 
\Bigl\{(x,z) \;\big|\; x_{\min} \leq x \leq x_{\max} \Bigr\}
\;\cap 
\Bigl\{(x,z) \;\big|\; z_{\min} \leq z \Bigr\} \\
&\;\setminus\; 
\Bigl\{(x,z) \;\big|\; x_{\mathrm{exc}}^{\min} \leq x \leq x_{\mathrm{exc}}^{\max}\Bigr\}
\end{aligned}
\end{equation}
where $(x_0,z_0)=(0,0)$,  
$r_{\mathrm{in}}=0.15\,\mathrm{m}$, $r_{\mathrm{out}}=0.35\,\mathrm{m}$,  
$x_{\min}=-0.25\,\mathrm{m}$, $x_{\max}=0.25\,\mathrm{m}$,  
$z_{\min}=-0.1\,\mathrm{m}$, and the exclusion zone  
$x_{\mathrm{exc}}^{\min}=-0.05\,\mathrm{m}$, $x_{\mathrm{exc}}^{\max}=0.05\,\mathrm{m}$.
Within this workspace, a narrow training region (highlighted in green in Fig.~\ref{fig:workspace}), covering less than 8\% of the total area, was selected based on the empirically derived feasible grasping workspace for the target object size, with sampling designed to ensure coverage and generalization within this feasible region \cite{wang2025spirobs}.

\subsubsection{Data Collection Strategy}
Control is realized in terms of tendon forces that drive the SpiRob’s motion. 
Expert demonstrations in simulation are generated via human teleoperation: an operator interactively controls the two tendons in real time (based on empirical experience and the grasping insights reported in~\cite{wang2025spirobs}). 
Concretely, during each rollout the operator adjusts the left/right tendon force commands to sequentially (i) uncurl and approach the target, and (ii) wrap and tighten to achieve enclosure, while observing the robot motion and contact state in the simulator.

For each expert simulated demo, we record the continuous control function $U_1^*(t)$, $t\in[0,T]$, representing the tendon forces over time, together with the corresponding object state $o\in\mathbb{R}^2$ denoting the sampled placement. 
The discrete control sequence used for training is obtained by applying the sampling operator $\mathcal{I}^{-1}[\cdot]$ to $U_1^*(t)$, yielding $U_1=\mathcal{I}^{-1}[U_1^*(t)]\in\mathbb{R}^{N\times 2}$, where $N$ is the number of uniformly sampled control steps over $[0,T]$. 
Fig.~\ref{fig:simulation_data} shows a representative teleoperated grasping demo in simulation, where the robot gradually approaches and encloses the object while $(U_1,o)$ are recorded for subsequent data processing.

\begin{figure}[t]
	\vspace*{0.3cm} 
    \centering
    \includegraphics[trim=0 0 0 0, clip, scale=1.25]{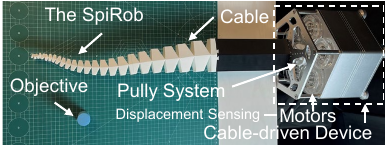}
    \caption{Experimental setup of the SpiRob platform.}
    \label{fig:experimental_setup}
\end{figure}

\subsection{Training and Inference Details}

We collected 30 expert demonstrations in MuJoCo, each providing a continuous expert control function $U_1^{*}(t) = [F_1^{*}, F_2^{*}$] with variable duration $T$ determined by the task completion time,where $F_1^{*}$ and $F_2^{*}$ denote the left and right tendon control forces.
This lightweight learning setup uses a small amount of training data, covering less than $8\% $ of the robot’s reachable workspace (Fig.~\ref{fig:workspace}).
To obtain fixed-length training data, the sampling operator $\mathcal{I}^{-1}[\cdot]$ was applied to each $U_1^{*}(t)$,  
yielding a discrete control sequence
$U_1 = [F_1, F_2] = \mathcal{I}^{-1}[U_1^{*}(t)] \in \mathbb{R}^{N \times 2}$, $N=80$ represents uniformly sampled timesteps across $[0,T]$.  
The model was trained for $50{,}000 $ iterations using the Adam optimizer~\cite{kingma2014adam}  
with a learning rate of $10^{-4}$ for Eq.~\ref{eqn:flow},  
on one NVIDIA GeForce RTX 4090 GPU, requiring approximately $6 $ hours to complete.  

During inference, the rectified flow model was sampled with $30$ iteration steps $\tau$  
(Fig.~\ref{fig:model_architecture}C), producing an inferred discrete expert-like control sequence $\hat{U}_1$,  
which is then interpolated using $\mathcal{I}[\cdot]$ to reconstruct the corresponding continuous expert-like control sequence  
$\hat{U}_1^{*}(t) = [\hat{F}_1^{*}, \hat{F}_2^{*}]=\mathcal{I}[\hat{U}_1]$  
for execution on the simulated robot.
\subsection{Simulation Validation}
Using the proposed training framework, only 30 demonstrations within a small region (less than 8\% of the reachable workspace) were used for training. The learned Rectified Flow policy was then evaluated on 1000 uniformly distributed targets across the full workspace,  
achieving a $97.5$\% grasping success rate in simulation (Fig.~\ref{fig:exp_overview}A). The six-layer Transformer, as shown in Fig. \ref{fig:model_architecture}, contains approximately 11 M parameters and requires $0.25$ s per sequence for inference. A lightweight single-layer variant reduces the parameter count to 0.7 M, trains in about one hour, and achieves a $94\%$ success rate.
\subsection{Comparison with Baselines in Simulation} To evaluate the
effectiveness of the proposed RF model, we compare it
with regression-based baselines for grasping tasks, including
MLP, CNN, and Transformer, under the same sparse training
setting of 10 demonstrations and identical training epochs.
As shown in Table \ref{tab:baseline}, across 100 test trials, our generative-
based method consistently outperforms all regression-based
approaches, demonstrating superior extrapolation, particularly in challenging working zones with limited expert data
coverage.

\begin{table}[!]
\caption{Success rate comparison with baselines in simulation.}
\centering
\begin{tabular}{ccccc}
\hline
Method & MLP  & CNN  & Transformer & Ours \\ \hline
Success rate     & 56\% & 68\% & 65\%        & 94\% \\ \hline

\end{tabular}
\label{tab:baseline}
\vspace{-5pt}
\end{table}

\section{Sim-to-real Experiments}\label{sec:experiments}

This section first introduces the experimental setup and control-signal processing for sim-to-real transfer,  
followed by validations in three scenarios: 
(i) workspace generalization from sparse demonstrations,  
(ii) geometric adaptability to object size variations, and  
(iii) robust dynamic scalability.
\subsection{Experimental Setup}
\begin{figure}[t]
	\vspace*{0.2cm} 
    \centering
    \includegraphics[trim=0 0 0 0, clip, scale=1]{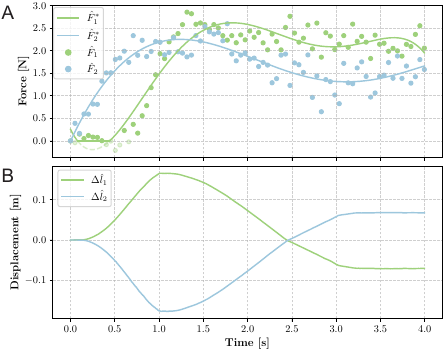}
    \caption{Control signal interpolation and tendon trajectory generation.}
    \label{fig:control_signal_processing}
\end{figure}

\begin{figure*}[t]
	\vspace*{0.2cm} 
    \centering
    \includegraphics[trim=0 0 0 0, clip, scale=0.95]{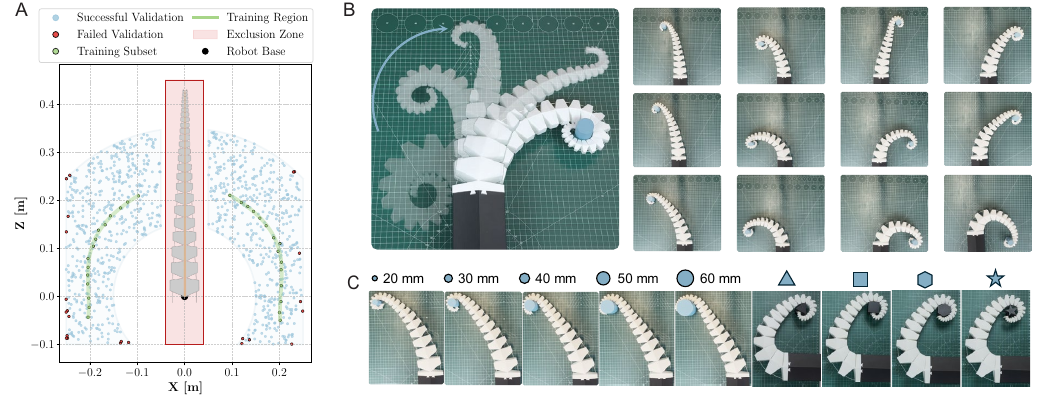}

    \caption{Sim-to-real results of workspace generalization from sparse demonstrations and geometric adaptability to object size variations}
    \label{fig:exp_overview}
\end{figure*}
\subsubsection{Hardware Platform}
The experimental platform is designed to closely mirror the MuJoCo simulation environment (Fig.~\ref{fig:experimental_setup}). 
The physical SpiRob has identical dimensions to the simulated model and is actuated through a cable driven device. 
This device is powered by high precision joint motors equipped with rotary encoders, allowing accurate control of tendon displacement and thus faithful execution of the commanded trajectories.
The robot is mounted on a tabletop with its motion plane aligned parallel to the surface, ensuring consistency with the simulated setup. 
\subsubsection{Force-to-Displacement Signal Processing for Sim-to-Real Transfer}
Bridging the learned policy from simulation to the physical platform requires a consistent representation of control inputs.  
In the robot dynamics formulation, tendon actuation appears as control-force terms in the equations of motion (Eq.~\eqref{soft_dynamics}), directly influencing the dynamic equilibrium of the soft robot.  
Accordingly, expert demonstrations are represented as continuous control-force functions $U_1^{*}(t)$,  
and the Rectified Flow model is trained to learn the corresponding force field  
$\hat{U}_1^{*}(t) = [\hat{F}_1^{*}, \hat{F}_2^{*}] = \mathcal{I}[\hat{U}_1]$.  
As shown in Fig.~\ref{fig:control_signal_processing}A, the discrete control sequence $\hat{U}_1= [\hat{F}_1, \hat{F}_2]$ is interpolated to produce a continuous force function $\hat{U}_1^{*}(t)$,  
ensuring physical consistency with the underlying dynamics.  
Here, tendon forces serve as direct control inputs, and the interpolated continuous function $\hat{U}_1^{*}(t)$ provides a physically consistent representation  
for subsequent signal conversion.

On the physical platform, tendon displacement control is adopted, as displacement control generally provides a more user-friendly and reliable interface than direct force control in robotic systems~\cite{craig2009introduction}.
For rigid robots, displacement control alone is often insufficient for highly dynamic and uncertain interaction tasks, since it cannot explicitly regulate contact forces.
In contrast, soft robots inherently exhibit embodied mechanical intelligence through compliance and redundancy, allowing adaptive and stable interactions to emerge even under displacement-based actuation.
As a result, direct control in the actuation space via tendon displacement offers both ease of implementation and the ability to preserve the intrinsic adaptability of soft bodies.

Accordingly, the hardware executes control through tendon displacement, with motor encoders providing accurate position feedback.
To ensure consistency between simulation and hardware execution, the inferred expert-like force trajectories $\hat{U}_1^{*}(t)$ are first applied in simulation offline to compute the corresponding tendon displacement profiles $\Delta \hat{l}(t)$, defined as
\begin{equation}
\small
\Delta \hat{l}_i(t) = l_i(q(t)) - l_i(q_0), \quad i=1,2,
\label{eq:tendon_displacement}
\end{equation}
where $l_i(q(t))$ denotes the tendon length at configuration $q(t)$ and $q_0=\mathbf{0}$ is the undeformed reference configuration. Note that trained at a nominal friction coefficient of 0.10, the RF model maintains successful execution under ±50\% friction mismatch in both simulation and hardware, indicating sufficient inherent generalization without domain randomization.
The resulting displacement profiles $\Delta \hat{l}(t)$ are then directly used as executable motor commands on the physical platform (Fig.~\ref{fig:control_signal_processing}B).

\subsection{Experimental Validation}

Based on the simulation results, 50 representative targets were uniformly selected across the workspace for real-world experiments to evaluate the SpiRob’s workspace coverage, geometric adaptability, and dynamic scalability.

\subsubsection{Workspace Generalization from Sparse Demonstrations}

To verify real-world performance, the 50 representative targets were tested on the physical SpiRob platform, achieving a $100\%$ success rate and confirming strong generalization and consistent sim-to-real transfer (Fig.~\ref{fig:exp_overview}B).

\subsubsection{Geometric Adaptability to Shape and Size Variations}

The Rectified Flow policy was trained exclusively on 30\,mm cylinders. During evaluation, the same policy was applied to diverse geometries without retraining: cylinders ranging from 20\,mm ($-33\%$) to 60\,mm ($+100\%$), as well as triangular, square, hexagonal, and star-shaped prisms. The SpiRob successfully completed all grasping tasks across these variations. These results demonstrate that the learned policy, combined with the robot's embodied mechanical intelligence, enables robust adaptation to both size variations ($-33\%$ to $+100\%$) and cross-shape generalization (Fig.~\ref{fig:exp_overview}C).

\subsubsection{Robust Dynamic Scalability}

\begin{figure}[t]
    \centering
    \includegraphics[trim=0 0 0 0, clip, scale=0.5]{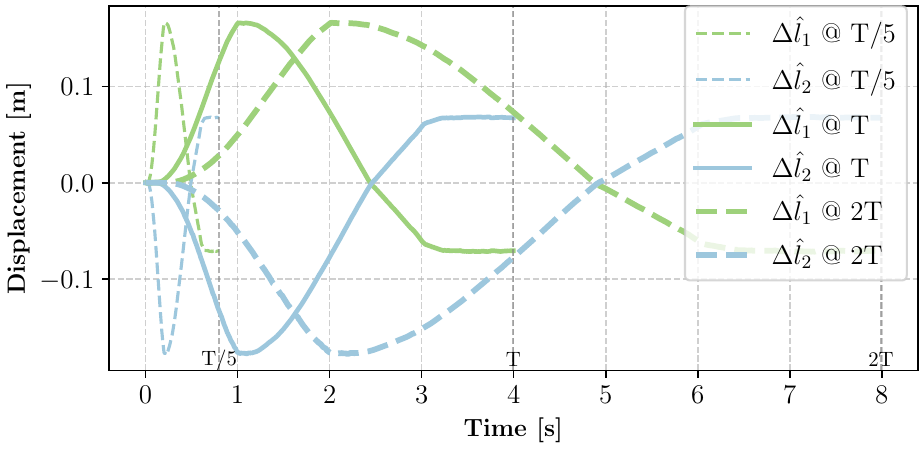}
    \caption{Experimental results of robust dynamic scalability}
    \label{fig:temporal_scaling}
\end{figure}

The policy’s adaptability was further tested under different execution speeds. Starting from a reference trajectory with duration $T$,  
the temporal axis was rescaled to generate faster or slower motions while maintaining the same displacement pattern (Fig.~\ref{fig:temporal_scaling}). The SpiRob achieved stable grasping performance for speeds ranging from 5× faster ($T/5$) to 2× slower ($2T$) than the nominal duration, demonstrating robust temporal scalability of the learned control pattern.

\section{Insights: Distributional Control and Actuation-Space Demonstration}

The experimental results in Section~\ref{sec:experiments} verify the proposed framework on a SpiRob, demonstrating the synergy between {embodied mechanical intelligence} and a {lightweight learning algorithm}.
These findings reveal key principles underlying {distributional control} and {actuation-space demonstration} in soft robotic manipulation.

Soft robots are inherently {underactuated}, and their configuration $q$ cannot be uniquely determined from actuator commands.
Although this property limits precise configuration-space control, it also provides {embodied mechanical intelligence}, allowing the robot to passively adapt to contact and environmental uncertainty through compliance.
Such embodiment reduces computational burden and enables the body itself to contribute to control.
Demonstrations are therefore more effective in the {actuation space}, where control inputs correspond directly to tendon forces (Fig.~\ref{comparison}B).
This representation preserves the physical mapping between actuation and deformation, allowing policies to leverage compliance for adaptive task execution.
In contrast, deterministic mappings in configuration space suppress the variability that facilitates adaptation. 

\begin{figure}[t]
    \centering
    \includegraphics[trim=0 0 0 0, clip, scale=0.6]{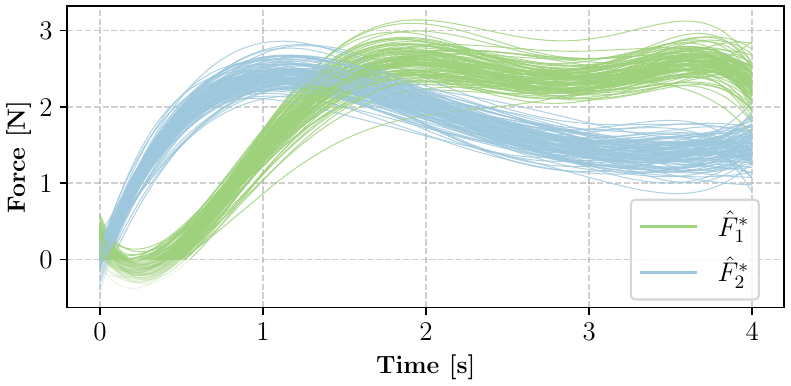}
    \caption{Distributional sampling of control sequences at one target position}
    \label{fig:distribution}
\end{figure}

Besides, the learned policy via the RF model exhibits a {distributional control pattern}. As shown in Fig.~\ref{fig:distribution}, 100 repeated inferences at one fixed target position produce force control functions that vary individually yet form a consistent distribution, with all executions converging to successful grasps. Rather than encoding a single trajectory, the policy represents a family of feasible control solutions, each providing a different realization through which the robot achieves the same task.
This distributional property enhances robustness to uncertainty.

\begin{figure}[t]
    \centering
    \includegraphics[trim=0 0 0 0, clip, scale=0.25]{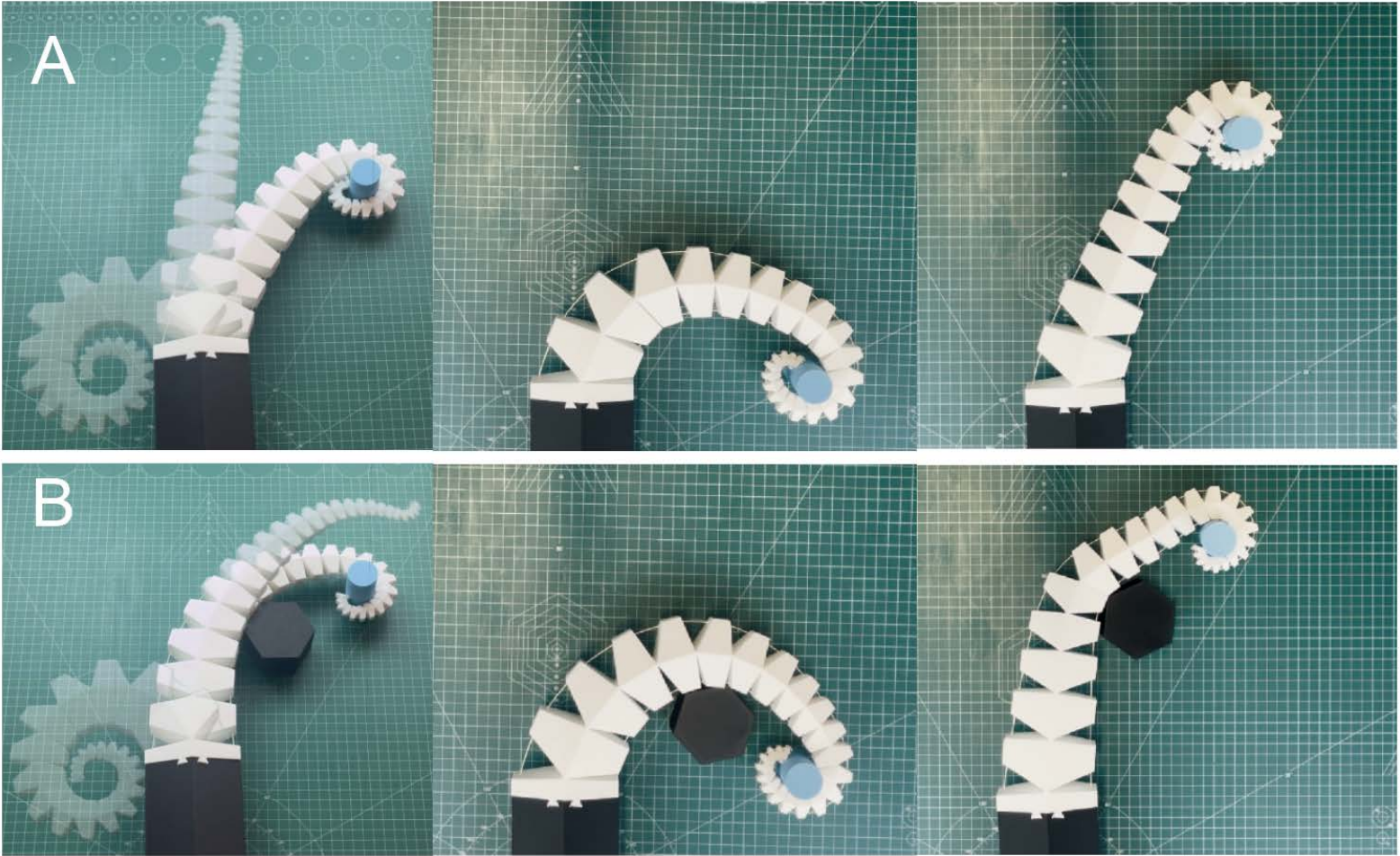}
    \caption{Whole-body grasping of a SpiRob in uncertain environments.}
    \label{fig:Whole-Body_interaction}
\end{figure}

As demonstrated in Fig.~\ref{fig:Whole-Body_interaction}, the SpiRob performs grasping both in free space (A) and near obstacles (B) using the same actuation-space distribution without explicit replanning or feedback correction.
The controller simply provides a feasible actuation distribution, while the compliant body autonomously accommodates contact and deformation in real time.
This synergy between {actuation-space representation} and {distributional control} embodies a broader principle of soft robotic intelligence.
It couples {embodied mechanical intelligence} with {distributional control policies}, {alleviating centralized computation} and enabling reliable whole-body grasping under uncertainty.

\section{conclusions}


This study presents a lightweight framework for learning from actuation-space demonstrations that couples embodied mechanical intelligence with distributional control to enable robust whole-body grasping in soft robots. Using a Rectified Flow model, the policy learns coordinated actuation patterns that exploit the soft body’s compliance and redundancy. Experiments on a SpiRob show data-efficient learning from few demonstrations, strong generalization to object variations, and robust performance under uncertainty, reducing reliance on precise centralized control.The evaluation of this paper focuses on a simplified yet contact-rich setup: targets are constrained (fixed) on a horizontal plane and parameterized by their 2D positions. While this design enables repeatable benchmarking and isolates the grasp-synthesis problem from uncontrolled object motion, it does not capture the full complexity of free-object, 3D, multi-contact manipulation.Future work will extend this framework to free-object and dynamic multi-contact manipulation, incorporate richer task suites (e.g., grasp-and-place), explore hierarchical policy structures, and investigate co-design methods~\cite{wang2023cuco} that jointly optimize morphology (e.g., spiral geometry and tendon routing) and control policies.

\bibliographystyle{IEEEtran}
\bibliography{reference}

\begin{thebibliography}{10}
\providecommand{\url}[1]{#1}
\csname url@samestyle\endcsname
\providecommand{\newblock}{\relax}
\providecommand{\bibinfo}[2]{#2}
\providecommand{\BIBentrySTDinterwordspacing}{\spaceskip=0pt\relax}
\providecommand{\BIBentryALTinterwordstretchfactor}{4}
\providecommand{\BIBentryALTinterwordspacing}{\spaceskip=\fontdimen2\font plus
\BIBentryALTinterwordstretchfactor\fontdimen3\font minus \fontdimen4\font\relax}
\providecommand{\BIBforeignlanguage}[2]{{%
\expandafter\ifx\csname l@#1\endcsname\relax
\typeout{** WARNING: IEEEtran.bst: No hyphenation pattern has been}%
\typeout{** loaded for the language `#1'. Using the pattern for}%
\typeout{** the default language instead.}%
\else
\language=\csname l@#1\endcsname
\fi
#2}}
\providecommand{\BIBdecl}{\relax}
\BIBdecl

\bibitem{mason2018toward}
M.~T. Mason, ``Toward robotic manipulation,'' \emph{Annual Review of Control, Robotics, and Autonomous Systems}, vol.~1, no.~1, pp. 1--28, 2018.

\bibitem{chu2023full}
H.~Chu, B.~J. Caasenbrood, M.~Keyvanara, I.~A. Kuling, and H.~Nijmeijer, ``Full-body grasping strategy for planar underactuated soft manipulators using passivity-based control,'' in \emph{2023 IEEE International Conference on Soft Robotics (RoboSoft)}.\hskip 1em plus 0.5em minus 0.4em\relax IEEE, 2023, pp. 1--7.

\bibitem{mengaldo2022concise}
G.~Mengaldo, F.~Renda, S.~L. Brunton, M.~B{\"a}cher, M.~Calisti, C.~Duriez, G.~S. Chirikjian, and C.~Laschi, ``A concise guide to modelling the physics of embodied intelligence in soft robotics,'' \emph{Nature Reviews Physics}, vol.~4, no.~9, pp. 595--610, 2022.

\bibitem{rao2021model}
P.~Rao, Q.~Peyron, S.~Lilge, and J.~Burgner-Kahrs, ``How to model tendon-driven continuum robots and benchmark modelling performance,'' \emph{Frontiers in Robotics and AI}, vol.~7, p. 630245, 2021.

\bibitem{wang2025spirobs}
Z.~Wang, N.~M. Freris, and X.~Wei, ``Spirobs: Logarithmic spiral-shaped robots for versatile grasping across scales,'' \emph{Device}, vol.~3, no.~4, 2025.

\bibitem{della2023model}
C.~Della~Santina, C.~Duriez, and D.~Rus, ``Model-based control of soft robots: A survey of the state of the art and open challenges,'' \emph{IEEE Control Systems Magazine}, vol.~43, no.~3, pp. 30--65, 2023.

\bibitem{chen2024data}
Z.~Chen, F.~Renda, A.~Le~Gall, L.~Mocellin, M.~Bernabei, T.~Dangel, G.~Ciuti, M.~Cianchetti, and C.~Stefanini, ``Data-driven methods applied to soft robot modeling and control: A review,'' \emph{IEEE Transactions on Automation Science and Engineering}, vol.~22, pp. 2241--2256, 2024.

\bibitem{chirikjian1992theory}
G.~S. Chirikjian, ``Theory and applications of hyper-redundant robotic manipulators,'' Ph.D. dissertation, California Institute of Technology, 1992.

\bibitem{tummers2023cosserat}
M.~Tummers, V.~Lebastard, F.~Boyer, J.~Troccaz, B.~Rosa, and M.~T. Chikhaoui, ``Cosserat rod modeling of continuum robots from newtonian and lagrangian perspectives,'' \emph{IEEE Transactions on Robotics}, vol.~39, no.~3, pp. 2360--2378, 2023.

\bibitem{duriez2013control}
C.~Duriez, ``Control of elastic soft robots based on real-time finite element method,'' in \emph{2013 IEEE international conference on robotics and automation}.\hskip 1em plus 0.5em minus 0.4em\relax IEEE, 2013, pp. 3982--3987.

\bibitem{chirikjian2015conformational}
G.~S. Chirikjian, ``Conformational modeling of continuum structures in robotics and structural biology: A review,'' \emph{Advanced Robotics}, vol.~29, no.~13, pp. 817--829, 2015.

\bibitem{george2018control}
T.~George~Thuruthel, Y.~Ansari, E.~Falotico, and C.~Laschi, ``Control strategies for soft robotic manipulators: A survey,'' \emph{Soft robotics}, vol.~5, no.~2, pp. 149--163, 2018.

\bibitem{giorelli2015neural}
M.~Giorelli, F.~Renda, M.~Calisti, A.~Arienti, G.~Ferri, and C.~Laschi, ``Neural network and jacobian method for solving the inverse statics of a cable-driven soft arm with nonconstant curvature,'' \emph{IEEE Transactions on Robotics}, vol.~31, no.~4, pp. 823--834, 2015.

\bibitem{thuruthel2018model}
T.~G. Thuruthel, E.~Falotico, F.~Renda, and C.~Laschi, ``Model-based reinforcement learning for closed-loop dynamic control of soft robotic manipulators,'' \emph{IEEE Transactions on Robotics}, vol.~35, no.~1, pp. 124--134, 2018.

\bibitem{gupta2016learning}
A.~Gupta, C.~Eppner, S.~Levine, and P.~Abbeel, ``Learning dexterous manipulation for a soft robotic hand from human demonstrations,'' in \emph{2016 IEEE/RSJ International Conference on Intelligent Robots and Systems (IROS)}.\hskip 1em plus 0.5em minus 0.4em\relax IEEE, 2016, pp. 3786--3793.

\bibitem{lecun2015deep}
Y.~LeCun, Y.~Bengio, and G.~Hinton, ``Deep learning,'' \emph{nature}, vol. 521, no. 7553, pp. 436--444, 2015.

\bibitem{ebert2021bridge}
F.~Ebert, Y.~Yang, K.~Schmeckpeper, B.~Bucher, G.~Georgakis, K.~Daniilidis, C.~Finn, and S.~Levine, ``Bridge data: Boosting generalization of robotic skills with cross-domain datasets,'' \emph{arXiv preprint arXiv:2109.13396}, 2021.

\bibitem{osa2018algorithmic}
T.~Osa, J.~Pajarinen, G.~Neumann, J.~A. Bagnell, P.~Abbeel, J.~Peters \emph{et~al.}, ``An algorithmic perspective on imitation learning,'' \emph{Foundations and Trends{\textregistered} in Robotics}, vol.~7, no. 1-2, pp. 1--179, 2018.

\bibitem{kober2013reinforcement}
J.~Kober, J.~A. Bagnell, and J.~Peters, ``Reinforcement learning in robotics: A survey,'' \emph{The International Journal of Robotics Research}, vol.~32, no.~11, pp. 1238--1274, 2013.

\bibitem{liu2023reinforcement}
X.~Liu, C.~D. Onal, and J.~Fu, ``Reinforcement learning of cpg-regulated locomotion controller for a soft snake robot,'' \emph{IEEE Transactions on Robotics}, vol.~39, no.~5, pp. 3382--3401, 2023.

\bibitem{satheeshbabu2019open}
S.~Satheeshbabu, N.~K. Uppalapati, G.~Chowdhary, and G.~Krishnan, ``Open loop position control of soft continuum arm using deep reinforcement learning,'' in \emph{2019 International Conference on Robotics and Automation (ICRA)}.\hskip 1em plus 0.5em minus 0.4em\relax IEEE, 2019, pp. 5133--5139.

\bibitem{alessi2024pushing}
C.~Alessi, D.~Bianchi, G.~Stano, M.~Cianchetti, and E.~Falotico, ``Pushing with soft robotic arms via deep reinforcement learning,'' \emph{Advanced Intelligent Systems}, vol.~6, no.~8, p. 2300899, 2024.

\bibitem{morimoto2022characterization}
R.~Morimoto, M.~Ikeda, R.~Niiyama, and Y.~Kuniyoshi, ``Characterization of continuum robot arms under reinforcement learning and derived improvements,'' \emph{Frontiers in Robotics and AI}, vol.~9, p. 895388, 2022.

\bibitem{nair2018overcoming}
A.~Nair, B.~McGrew, M.~Andrychowicz, W.~Zaremba, and P.~Abbeel, ``Overcoming exploration in reinforcement learning with demonstrations,'' in \emph{2018 IEEE international conference on robotics and automation (ICRA)}.\hskip 1em plus 0.5em minus 0.4em\relax IEEE, 2018, pp. 6292--6299.

\bibitem{chi2025diffusion}
C.~Chi, Z.~Xu, S.~Feng, E.~Cousineau, Y.~Du, B.~Burchfiel, R.~Tedrake, and S.~Song, ``Diffusion policy: Visuomotor policy learning via action diffusion,'' \emph{The International Journal of Robotics Research}, vol.~44, no. 10-11, pp. 1684--1704, 2025.

\bibitem{wu2025device}
Y.~Wu, H.~Wang, Z.~Chen, J.~Pang, and D.~Xu, ``On-device diffusion transformer policy for efficient robot manipulation,'' \emph{arXiv preprint arXiv:2508.00697}, 2025.

\bibitem{ma2024hierarchical}
X.~Ma, S.~Patidar, I.~Haughton, and S.~James, ``Hierarchical diffusion policy for kinematics-aware multi-task robotic manipulation,'' in \emph{Proceedings of the IEEE/CVF Conference on Computer Vision and Pattern Recognition}, 2024, pp. 18\,081--18\,090.

\bibitem{ze20243d}
Y.~Ze, G.~Zhang, K.~Zhang, C.~Hu, M.~Wang, and H.~Xu, ``3d diffusion policy: Generalizable visuomotor policy learning via simple 3d representations,'' \emph{arXiv preprint arXiv:2403.03954}, 2024.

\bibitem{rflow}
X.~Liu, C.~Gong, and Q.~Liu, ``Flow straight and fast: Learning to generate and transfer data with rectified flow,'' \emph{International Conference on Learning Representations (ICLR)}, 2023.

\bibitem{flowmatching}
Y.~Lipman, R.~T. Chen, H.~Ben-Hamu, M.~Nickel, and M.~Le, ``Flow matching for generative modeling,'' \emph{arXiv preprint arXiv:2210.02747}, 2022.

\bibitem{wang2024exploiting}
Z.~Wang and N.~M. Freris, ``Exploiting frictional effects to reproduce octopus-like reaching movements with a cable-driven spiral robot,'' in \emph{2024 IEEE 7th International Conference on Soft Robotics (RoboSoft)}.\hskip 1em plus 0.5em minus 0.4em\relax IEEE, 2024, pp. 537--542.

\bibitem{ddpm}
J.~Ho, A.~Jain, and P.~Abbeel, ``Denoising diffusion probabilistic models,'' \emph{Advances in neural information processing systems}, vol.~33, pp. 6840--6851, 2020.

\bibitem{howell1996loop}
L.~L. Howell and A.~Midha, ``A loop-closure theory for the analysis and synthesis of compliant mechanisms,'' \emph{Journal of Mechanical Design}, vol. 118, no.~1, pp. 121--125, 1996.

\bibitem{brockett1983asymptotic}
R.~W. Brockett \emph{et~al.}, ``Asymptotic stability and feedback stabilization,'' \emph{Differential geometric control theory}, vol.~27, no.~1, pp. 181--191, 1983.

\bibitem{dit}
W.~Peebles and S.~Xie, ``Scalable diffusion models with transformers,'' 2023.

\bibitem{attention}
A.~Vaswani, N.~Shazeer, N.~Parmar, J.~Uszkoreit, L.~Jones, A.~N. Gomez, {\L}.~Kaiser, and I.~Polosukhin, ``Attention is all you need,'' \emph{Advances in neural information processing systems}, vol.~30, 2017.

\bibitem{kingma2014adam}
D.~P. Kingma, ``Adam: A method for stochastic optimization,'' \emph{arXiv preprint arXiv:1412.6980}, 2014.

\bibitem{craig2009introduction}
J.~J. Craig, \emph{Introduction to robotics: mechanics and control, 3/E}.\hskip 1em plus 0.5em minus 0.4em\relax Pearson Education India, 2009.

\bibitem{wang2023cuco}
Y.~Wang, S.~Wu, H.~Fu, Q.~Fu, T.~Zhang, Y.~Chang, and X.~Wang, ``Curriculum-based co-design of morphology and control of voxel-based soft robots,'' in \emph{International Conference on Learning Representations (ICLR)}, 2023.

\end{thebibliography}

\end{document}